\documentclass[mnsc,nonblindrev]{informs3_homepage}

\OneAndAHalfSpacedXI

\usepackage{macros}

 \usepackage{natbib}
\bibpunct[, ]{(}{)}{,}{a}{}{,}%
\def\bibfont{\small}%

\usepackage[utf8]{inputenc} 
\usepackage[T1]{fontenc}    
\usepackage{booktabs}       
\usepackage{amsfonts}       
\usepackage{nicefrac}       
\usepackage{microtype}      
\usepackage{xcolor}         
\usepackage{graphicx}
\usepackage{babel}
\usepackage{amsmath}
\usepackage{tikz}

\usepackage{multirow}
\usepackage{hhline}
\usepackage{hyperref}       
\usepackage{url}            
\usetikzlibrary{matrix,decorations.pathreplacing,calc,positioning}

\begin{document}

	\RUNAUTHOR{}
	\RUNTITLE{Post Launch Evaluation of Policies}
    \TITLE{Post Launch Evaluation of Policies in a High-Dimensional Setting}
	
\ARTICLEAUTHORS{
    \AUTHOR{Shima Nassiri}
    \AFF{Lyft\textsuperscript{*}}
    \AUTHOR{Mohsen Bayati}
    \AFF{Amazon and Stanford University\textsuperscript{*}}
    \AUTHOR{Joe Cooprider}
    \AFF{Amazon}
}
\renewcommand{\thefootnote}{*}
\footnotetext{Work completed while at Amazon.}
	
	\ABSTRACT{
A/B tests, also known as randomized controlled experiments (RCTs), are the gold standard for evaluating the impact of new policies, products, or decisions (collectively referred to as ``treatments''). However, these tests can be costly in terms of time and resources, potentially exposing users, customers, or other test subjects (collectively called ``units'') to inferior options.
This paper explores practical considerations in applying methodologies inspired by ``synthetic control'' as an alternative to traditional A/B testing in settings with very large numbers of units, involving up to hundreds of millions of units, which is common in modern applications such as e-commerce and ride-sharing platforms. This method is particularly valuable in settings where the treatment affects only a subset of units, leaving many units unaffected. In these scenarios, synthetic control methods leverage data from unaffected units to estimate counterfactual outcomes for treated units. After the treatment is implemented, these estimates can be compared to actual outcomes to measure the treatment effect. A key challenge in creating accurate counterfactual outcomes is interpolation bias, a well-documented phenomenon that occurs when control units differ significantly from treated units. To address this, we propose a two-phase approach: first using nearest neighbor matching based on unit covariates to select similar control units, then applying supervised learning methods suitable for high-dimensional data to estimate counterfactual outcomes. Testing using six large-scale experiments demonstrates that this approach successfully improves estimate accuracy. However, our analysis reveals that machine learning bias---which arises from methods that trade off bias for variance reduction---can impact results and affect conclusions about treatment effects. We document this bias in large-scale experimental settings and propose effective de-biasing techniques to address this challenge.}
		
	\maketitle

     \section{Introduction}\label{sec:intro}

Modern technology companies rely on experimentation to drive their decision-making processes. Each year, major technology companies conduct thousands of controlled experiments to evaluate new features, products, algorithms, and policy changes before launching them at scale \citep{gupta2019top}. These experiments, commonly known as A/B tests, have become the standard method for assessing potential changes across diverse business functions and engineering decisions.

The design of an A/B test allows for a thorough isolation of the treatment effect, minimizing the influence of other concurrent treatments, unit differences and temporal trends on the results \citep{athey2016economics,kohavi2007practical}. This precision is achieved through forming a control group that is comparable to the treatment group. The comparability is ensured through careful randomization that distributes the units between the control and treatment groups such that the treatment groups are direcly comparable. Additionally, an A/B test allows for real and live responses to a new treatment, and hence the treatment outcomes after launching the treatment.

A/B tests, while effective, involve significant costs in both time and resources. During an experimental period, units may be exposed to suboptimal treatments, either through the direct implementation of an inferior option or through opportunity costs when delaying the rollout of a superior treatment. The challenges extend beyond the initial testing phase. Once a treatment is launched, evaluating whether its benefits persist beyond the initial experimental period requires additional experimental efforts. Traditional approaches include designing new A/B tests that temporarily suspend the treatment for a subset of units or maintaining a small holdout group for ongoing comparison. These monitoring methods present their own difficulties: they can be expensive to maintain, may yield imprecise results due to limited sample sizes, and often become unreliable over time as other changes and treatments affect the control or treatment population.

Further, several important scenarios call for alternative methods of measuring treatment impact outside of traditional A/B testing. These situations arise when immediate implementation is required due to urgent business needs or other constraints, making controlled experiments impractical. They also occur when the environment features multiple concurrent changes that could affect treatment effectiveness, requiring continuous evaluation. Alternative measurement approaches become particularly valuable when considering whether to discontinue existing treatments, as running new experiments would be resource-intensive, or when the costs of conducting full-scale A/B tests exceed the anticipated benefits of the treatment itself. In these circumstances, one needs a reliable and cost-effective methodology to assess treatment impact after implementation.

To address these scenarios, we can leverage the fact that there are numerous units unaffected or minimally impacted by the treatment that fall outside the treated population for a given experiment (referred to as the \emph{donor pool}). This donor pool captures trends not associated with the treatment and helps us predict outcomes as if a control group were present. Consequently, we can construct \emph{counterfactual control outcomes}, which can then be compared to the affected units after launch to estimate the treatment's impact.

Generating counterfactual control outcomes is well-studied in the literature, with foundational work by \cite{abadie2003economic} and \cite{abadie2010synthetic}, followed by substantial subsequent research. Under these frameworks, outcomes of units are measured over time (panel data), and a donor set is considered to predict what would have happened to a treated unit in the absence of intervention in the post-intervention period, thereby building the counterfactual control outcomes. Counterfactual outcomes for the control group of each treated unit are estimated using convex regression or, more recently, machine learning techniques \citep{imbens2017balancing,amjad2018robust,athey2021matrix,agarwal2020synthetic}; see \citep{arkhangelsky2023causal} and references therein for further details. However, this approach can be compromised by two sources of bias. The first, \emph{interpolation bias}, occurs when the donor pool units differ significantly from the units under intervention \citep{abadie2015comparative} even after controlling for covariates. The second, more nuanced source is \emph{machine learning bias}, which could be potentially introduced by methods that trade off bias for variance reduction as they are trained to predict accurate pre-treatment values for the treated units. A simple example is Lasso \citep{chen1995examples,tibshirani1996regression}, which introduces bias to the coefficient estimates in order to reduce variance of the predictions. This bias is similar to the ones studied recently by \cite{bottmer2024designbased}, and is particularly prevalent in extremely high-dimensional settings involving large donor pools (with up to hundreds of millions of units) and highly noisy outcome data, leading to substantial bias. Such high-dimensional settings are increasingly common in modern technology applications---for instance, in online ticket marketplaces where individual transactions serve as units \citep{blake2021price}.

In this paper, we propose a two-phase approach to address these challenges. In the first phase, we use an efficient implementation of approximate nearest neighbor matching based on unit covariates to filter the donor pool to a subset of units with characteristics similar to the treated units, targeting the interpolation bias. Due to the substantially large donor size (often on the order of hundreds of millions), this matching phase also serves the practical necessity of subsampling. In the second phase, we utilize supervised learning methods for extremely high-dimensional data, building on techniques from \citep{imbens2017balancing}, \citep{amjad2018robust}, and \citep{agarwal2020synthetic}. Testing this approach on data from six randomized experiments demonstrates its effectiveness in reducing interpolation bias. However, our analysis reveals that machine learning bias continues impacts results, potentially affecting conclusions about treatment effects. We document this bias affecting treatment effect estimation in real experimental settings and introduce de-biasing techniques that effectively address this challenge.

     \section{Setup and Methods}\label{sec:methods}

In this section, we introduce notation for representing control and treated outcomes and define the goal of estimating treatment effects. We describe the task of predicting counterfactuals for treated units and outline methods for estimating both average (ATE) and heterogeneous treatment effects (HTE). 

Consider the (panel) data matrix $\mY(0)$ representing the untreated outcomes:
\vspace{5mm}

\begin{center}
	\begin{tikzpicture}
		\matrix [matrix of math nodes,left delimiter=(,right delimiter=),row sep=0.05cm,column sep=0.05cm] (m) {
			Y_{11}(0)&Y_{12}(0)&\ldots&Y_{1T_0}(0)&?&\ldots&? \\
			\vdots&\vdots&&\vdots&&&\\
			Y_{i1}(0)&Y_{i2}(0)&\ldots&Y_{iT_0}(0)&?&\ldots&? \\
			\vdots&\vdots&&\vdots&&&\\
			Y_{n1}(0)&Y_{n2}(0)&\ldots&Y_{nT_0}(0)&?&\ldots&? \\
			Y_{(n+1)1}(0)&Y_{(n+1)2}(0)&\ldots&Y_{(n+1)T_0}(0)&Y_{(n+1)(T_0+1)}(0)&\ldots&Y_{1T}(0) \\
			\vdots&\vdots&&\vdots&&&\vdots \\
			Y_{N1}(0)&Y_{N2}(0)&\ldots&Y_{NT_0}(0)&Y_{N(T_0+1)}(0)&\ldots&Y_{NT}(0)\\};
		\draw[dashed] (m-6-1.north west) -- (m-6-7.north east);
		\draw[dashed] ([xshift=.3cm]m-1-4.north east) -- ([xshift=.3cm]m-8-4.south east);
		\node[above=3pt of m-1-2] (top-2) {Pre-treatment};
		\node[above=3pt of m-1-5] (top-5) {Post-treatment};
		\draw[decorate,decoration={brace,amplitude=10pt,mirror,raise=5pt}] ([xshift=-.7cm]m-1-1.west) -- ([xshift=-.7cm]m-5-1.west) node [black,midway,xshift=-.8cm,rotate=90] {\footnotesize Treated units};
		\draw[decorate,decoration={brace,amplitude=10pt,mirror,raise=5pt}] ([xshift=-.5cm]m-6-1.north west) -- ([xshift=-.7cm]m-8-1.south west) node [black,midway,xshift=-.8cm,rotate=90] {\footnotesize Donor units};
		\node[left=2cm of m-5-1] (textnode) {$\mY(0) =$};
	\end{tikzpicture}
\end{center}
Here, rows represent units and columns represent time periods. There are $N$ total units, of which $n$ are treated, while the remaining $N-n$ units are untreated and serve as donor units. Treatment is applied to the treated units starting from time $T_0+1$. Let $[m]$ denote the set $\{1, 2, \ldots, m\}$ for any integer $m$. Using the potential outcomes framework, for each $i \in [N]$ and $t \in [T]$, the untreated outcome of unit $i$ in period $t$ is denoted by $Y_{it}(0)$, and the treated outcome is denoted by $Y_{it}(1)$. We observe $Y_{it}(0)$ for all units and time periods, except for $i \in [n]$ and $t > T_0$, where we only observe the treated values $Y_{it}(1)$. The treatment assignment is indicated by $W_{it}$, where $W_{it} \in \{0, 1\}$. Specifically, $W_{it}=0$ for all $i$ and $t$ except when $i \in [n]$ and $t > T_0$, in which case $W_{it}=1$. The goal is to estimate the \emph{average treatment effect} (ATE),
\begin{equation}\label{eq:ATE}
	\tau = \frac{\sum_{i \in [N]} \sum_{t \in [T]} W_{it} \left[ Y_{it}(1) - Y_{it}(0) \right]}{\sum_{i \in [N]} \sum_{t \in [T]} W_{it}}.
\end{equation}

Following the synthetic control literature pioneered by \cite{abadie2003economic}, we form predictions $\hat{Y}_{it}(0)$ for the counterfactual values $Y_{it}(0)$ in the treated pairs $(i,t)$ and estimate $\tau$ as

\begin{equation}\label{eq:estimateATE}
	\hat{\tau} = \frac{\sum_{i \in [N]} \sum_{t \in [T]} W_{it} \left[ Y_{it}(1) - \hat{Y}_{it}(0) \right]}{\sum_{i \in [N]} \sum_{t \in [T]} W_{it}}\,.
\end{equation}

\paragraph{Heterogeneous Treatment Effects (HTE).} Estimating counterfactual values for each pair $(i,t)$ allows us to calculate treatment effects at the individual unit and time-period levels. Specifically, we denote the HTE for unit $i$ at time $t$ by $\tau_{it}$, where $\tau_{it} = Y_{it}(1) - Y_{it}(0)$. We estimate $\tau_{it}$ using $\hat{\tau}_{it} = Y_{it}(1) - \hat{Y}_{it}(0)$. In Section \ref{sec:HTE}, we will explore this in more detail.

\subsection{Counterfactual Prediction}

To obtain the counterfactual predictions, we can frame the problem as a multi-label supervised learning (regression) task, where the columns of $\mY$ represent observations. The first $T_0$ columns serve as the training data, while the remaining columns represent the test data. This approach is known as \emph{vertical regression} in the causal inference literature \citep{athey2021matrix}, as the observations are organized by columns. The first $n$ rows of $\mY$ serve as labels, and the remaining rows are treated as features. Additionally, for each unit $i$, we have access to a covariate vector $\vx_i \in \mathbb{R}^p$, representing the unit-specific features. While these covariates do not naturally integrate into the supervised learning task described above, they are utilized in the two-phase method outlined below.

In our experiments, the number of units $N$ ranges from hundreds of thousands to hundreds of millions, depending on the experiment and treatment type, with $n$ being a constant fraction of $N$. On the other hand, $T$ (the number of time periods) is relatively small, typically less than 100. Thus, the supervised learning task is extremely high-dimensional.

At first glance, one might consider transposing the above supervised learning task (referred to as horizontal regression in the causal inference literature \citep{athey2021matrix}). This approach removes the dimensionality problem since $N \gg T$, and it also allows the covariate vectors $\vx_i$ to be incorporated as additional features. In this setup, future outcomes are predicted using pre-treatment outcomes. However, due to the highly non-stationary nature of the data (i.e., significant variation over time), we found this approach to perform poorly in practice. Furthermore, matrix completion methods, such as those proposed by \cite{athey2021matrix} or \cite{farias2021learning}, are infeasible here due to the extremely large size of $N$. In fact, as noted in \cite{athey2021matrix}, in settings where $N \gg T$, their matrix completion method performs similarly to horizontal regression.

Here, we go through a variety of candidate vertical regression based supervised learning approaches. Note that given the large values of $N$ (and $n$) in some applications, we often need to subsample the donor set (e.g., taking 1\% of the donors) to make these benchmarks computationally feasible.

\paragraph{Nearest Neighbor Baseline (NN).} For each treated unit $i$, we can find the closest donor unit $d_i$ based on a distance metric applied to the pre-treatment outcomes (the first $T_0$ columns of rows $i$ and $d_i$). We define the counterfactual estimate as $\hat{Y}_{it}(0) = Y_{d_it}(0)$. This method can be extended to k-NN, where the counterfactual is the average of the $k$ nearest donors' outcomes.

\paragraph{Ridge Regression Baseline (Ridge).} For each treated unit $i$, we apply ridge regression to the high-dimensional multi-label (multi-task) regression problem. Lasso or ElasticNet can also be used in place of Ridge, but we observed similar performance across these methods, so we only report results for Ridge.

\paragraph{Principal Component Regression Baseline (PCR).} We construct a set of $k$ ($k \leq T$) vectors that span the row space of the submatrix of $\mY(0)$ corresponding to the donor units. Then, for each treated unit $i \in [n]$, we regress the $i$-th row of $\mY$ onto these $k$ ``latent'' donors. Specifically, if the singular value decomposition (SVD) of this submatrix is $U\Sigma V^\top$, the matrix $V^\top$ will have $T$ columns and up to $T$ rows. The first $k$ rows of $V^\top$ correspond to the latent donors.

This method, followed by an appropriate selection of the rank $k$, is central to the approach of \cite{agarwal2020synthetic}, which uses a hard thresholding criterion as described in \cite{gavish2014optimal}. 

\paragraph{PCRRidge and PCRLasso.} In this variant of PCR, instead of hard-thresholding to select a smaller rank $k$, we use all latent donors and apply ridge regression or Lasso to estimate the counterfactuals.

\paragraph{Other Approaches.} We also experimented with other baselines, such as replacing Ridge and Lasso with robust regression, random forest, boosted trees, kernel ridge regression, or multi-layer neural networks. However, due to the high-dimensional, low-sample nature of the problem, these methods performed poorly and were outperformed by the previously mentioned benchmarks. As a result, their outcomes are not detailed in the final results. Furthermore, these methods were significantly slower, particularly given the large value of $N$.

\subsection{Two-phase Prediction}

Motivated by the discussions in \cite{abadie2015comparative}, we begin our analysis by filtering the donor set by selecting those most similar to the treated units. Specifically, for each treated unit $i$, we select its $k$ nearest neighbors based on the distance between their covariate vectors $\vx$. The rows of $\mY$ corresponding to the remaining donors are then removed from $\mY$. After this first phase, we apply one of the vertical regression baselines (e.g., k-NN, PCR, Ridge, PCRRidge, or PCRLasso). We refer to these two-phase variants as k-NN-2, PCR-2, Ridge-2, PCRRidge-2, or PCRLasso-2, respectively.

\paragraph{Computational Benefits.} In addition to reducing the bias of counterfactual predictions that can arise from ``far away'' donors, filtering the donor set decreases $N$ by an order of magnitude or more, which substantially accelerates the second phase. As mentioned earlier, in single-phase vertical regression, we often need to randomly subsample the donor set to make the task manageable, especially when methods like Lasso or Ridge require hyperparameter tuning.

For the first-stage $k$-NN, we used the Annoy (Approximate Nearest Neighbors Oh Yeah) package\footnote{\url{https://github.com/spotify/annoy}}, which offers an efficient implementation for approximating $k$-NN. For $k$ values below 20, this stage typically completes in under 10 minutes, even for $N$ and $n$ values in the hundreds of millions, on a standard computer.

\paragraph{Spillovers.} A major concern in randomized experiments is spillover from treated units to control units, commonly referred to as \emph{network interference} in the causal inference or experimental design literature. This occurs when the Stable Unit Treatment Value Assumption (SUTVA) \citep{cox1958planning, rubin1978bayesian} is violated. In our context, spillovers can bias our results if the donor set is not carefully curated. Specifically, if the donor set includes units that are influenced by the treatment, the resulting estimates will be biased. To mitigate this, we exclude donor units that have strong connections to treated units from the donor pool.

\subsection{Prediction Accuracy and Model Selection}
\label{subsec:model-selection}

Although our primary goal is to estimate the ATE, $\tau$, we use an interim metric to assess model performance and guide model selection. Specifically, we evaluate the relative error between the predictions and the actual values, which is defined as:
\begin{equation}
	\text{relative error} = \frac{\|\text{Prediction} - \text{Actual}\|}{\|\text{Actual}\|}\,,\label{eq:relative-error}
\end{equation}
where ``Actual'' refers to the sub-matrix of $\mY(0)$ corresponding to the treated (unit, time) pairs, and ``Prediction'' refers to their predicted values. For the norm, we experimented with both the Frobenius norm and the $\ell_1$ norm of the matrices. Specifically, for a matrix $\mA$, the Frobenius norm is defined as $\sqrt{\sum_{it}A_{it}^2}$, while the $\ell_1$ norm is defined as $\sum_{it}|A_{it}|$. We found that the $\ell_1$ norm yielded more reliable performance as it is less sensitive to extreme values in the outcome distribution. This robustness is particularly important in settings where the outcome data exhibits high variance and contains significant outliers. The Frobenius norm, by squaring the differences, can disproportionately amplify the impact of these outliers on model selection, potentially leading to suboptimal choices of prediction methods and hyperparameters.

     \section{Empirical Results and Bias Correction}\label{sec:empirics}

A major challenge in evaluating the performance of ATE estimation methods in practice is that they are often applied in observational settings, where the true ATE is unknown. Without access to the ground truth, it is difficult to determine whether the methods are producing accurate estimates. To address this, we leveraged data from six large-scale historical A/B test experiments, where an unbiased estimate of the true ATE is known. This unique setup allowed us to directly compare the estimates generated by the methods described in Section \ref{sec:methods} with the actual ATE values. Our analysis consists of two main parts. First, we demonstrate the benefits of the two-stage approach over the single-stage approach, specifically examining how the nearest-neighbor matching in the first stage impacts the donor pool's distribution, prediction accuracy of counterfactual outcomes, and ability to recover the ground truth ATE. We then conduct a more comprehensive assessment of the two-stage solution, evaluating its capability to replicate experimental results, examining the impact of machine learning bias, and demonstrating how we address this bias.

\subsection{Benefit of the Two-Phase Approach}

In this section, we compare the two-phase solution with the single-phase approach. Due to the massive scale of the donor pool, even the single-phase approach requires subsampling. We use a 1\% subsample of the donor pool, bringing its size approximately equal to that obtained by nearest neighbor matching in the two-phase approach.

\paragraph{Donor pool alignment.} Table \ref{tab:distribution-comparison} illustrates how our two-stage approach substantially improves the alignment between the donor pool and experimental groups' outcome distributions. In the single-phase approach, the donor pool exhibits markedly different characteristics, particularly at the distribution tails, with the 1st percentile at -18.9 (compared to -0.7 for control) and the 99th percentile at 47.7 (compared to 22.6 for control). The two-phase approach significantly reduces these disparities, bringing the donor pool's extreme values (-2.3 at the 1st percentile and 32.2 at the 99th percentile) much closer to those of the experimental groups. This improved alignment in the outcome distributions provides strong evidence that our nearest-neighbor matching effectively selects a more representative donor pool for counterfactual prediction.
\begin{table}[!htbp]
\centering
\caption{Comparison of outcome distributions across experimental and potential donor units. The table shows key quantiles of the outcome distribution for three groups: the pool of donor units available, the experimental control group, and the treatment group. Values are shown for both single-phase and two-phase approaches to demonstrate the improved alignment achieved through nearest-neighbor matching.}
\label{tab:distribution-comparison}
\begin{tabular}[t]{lccccccl}
\toprule
& \multicolumn{6}{c}{Quantile} & \\
\cmidrule(lr){2-7}
Group & 0.01 & 0.05 & 0.10 & 0.90 & 0.95 & 0.99 & Approach \\
\midrule
Donor Units & -18.9 & 0.0 & 0.0 & 6.4 & 8.3 & 47.7 & Single-phase \\
Experimental Control & -0.7 & 0.0 & 0.2 & 6.3 & 7.2 & 22.6 & Single-phase \\
Experimental Treatment & -0.4 & 0.0 & 0.1 & 6.3 & 7.3 & 24.0 & Single-phase \\
\midrule
Donor Units & -2.3 & 0.0 & 0.1 & 6.5 & 8.0 & 32.2 & Two-phase \\
Experimental Control & -0.7 & 0.0 & 0.2 & 6.3 & 7.2 & 22.6 & Two-phase \\
Experimental Treatment & -0.4 & 0.0 & 0.1 & 6.3 & 7.3 & 24.0 & Two-phase \\
\bottomrule
\end{tabular}
\end{table}

\paragraph{Accuracy of counter-factual prediction.} Since the control units in each A/B test experiment are untreated, we can use their observed outcomes as the ground truth we want to predict in Eq. \eqref{eq:relative-error}. Using this approach across all six experiments, we observe that the performance of prediction models improves by about $10\%$ when using a two-phase approach compared to a single-phase approach.

\paragraph{Replicating the experiment.} We can apply the methodology from Section \ref{sec:methods} to the predicted counterfactuals of the treated units in any historical A/B test experiment and estimate the ATE (i.e., ignoring the control group). Alternatively, we can calculate the ground truth ATE by comparing the treated units' outcomes with the control units' outcomes. Similar to standard experimental design terminology, where A/B tests measure treatment effects and A/A tests verify null effects, we perform two validations. The suffix ``-ST'' (Synthetic Treatment) in our notation indicates that we are using synthetically generated counterfactuals for the treated units.

\textbf{Validation (A/B-ST):} This validation mirrors the purpose of traditional A/B tests. We compare:
\begin{itemize}
\item Ground truth: ATE from the actual experiment (A/B).
\item Estimate: ATE by comparing outcomes for treated units with their synthetic (predicted) counterfactual control outcomes (A/B-ST).
\end{itemize}
A successful validation confirms our ability to detect genuine treatment effects, showing alignment between synthetic and actual estimates when true effects exist.

\textbf{Validation (A/A-ST):} This validation mirrors the purpose of traditional A/A tests. We compare
\begin{itemize}
\item Ground truth: Zero (no effect).
\item Estimate: ATE obtained by comparing the observed control unit outcomes with the counterfactual control outcomes for the treated unit.
\end{itemize}
A successful validation confirms we do not falsely detect effects, showing no significant differences between synthetic and actual control outcomes.

Next, we apply these two validations to both the single-phase and two-phase methods for one of the six experiments and the estimation method PCRRidge. As shown in Table \ref{Tab:validation_KNN_PCRR_E1}, we observe a failed validation (A/B-ST) under the single-phase approach. We observe that for A/B-ST, a false effect is detected in this case, while the ground truth did not indicate a significant effect. On the other hand, when using the two-phase approach, we observe that the ATE from the counterfactual control produced by PCRRidge-2 is aligned with that of the observed control in terms of direction and statistical significance. Additionally, the second validation passes as the A/A-ST test does not indicate a significant difference between the counterfactual and observed control outcomes.
\begin{table}[!htbp]
\centering
\caption{Validation results comparing treatment effect estimates between experimental ground truth and counterfactual predictions. The table shows the estimated treatment effect ($\tau$), its statistical significance (P-value), and validation outcomes for both single-phase and two-phase approaches using the PCRRidge method.}
\label{Tab:validation_KNN_PCRR_E1}
\begin{tabular}[t]{llcccc}
\toprule
Method & Label & $\tau$ & P-value & Validation & Approach \\
\midrule
Ground truth & A/B & -0.14 & 0.47 & -- & -- \\
PCRRidge & A/B-ST & -0.42 & 0.01 & {\color{pastelred}Fail} & Single-phase \\
PCRRidge & A/A-ST & 0.28 & 0.14 & {\color{pastelgreen}Pass} & Single-phase \\
PCRRidge-2 & A/B-ST & -0.28 & 0.1 & {\color{pastelgreen}Pass} & Two-phase \\
PCRRidge-2 & A/A-ST & 0.13 & 0.47 & {\color{pastelgreen}Pass} & Two-phase \\
\bottomrule
\end{tabular}
\end{table}

\subsection{Assessment of the Two-Phase Approach Across Six Experiments}

We next apply the above two validations (A/B-ST) and (A/A-ST) for the two-phase approach across six other historical experiments, where the estimation method is selected via cross-validation during training. Table \ref{Tab:validation_histlabs} summarizes the results for the ground truth and estimated ATE for these validations.
\begin{table}[!htbp]
\centering
\caption{Comprehensive validation of the two-phase approach across six historical experiments. The table compares three key metrics: the ground truth treatment effect from actual A/B tests (A/B), the estimated effect using counterfactual predictions (A/B-ST), and the placebo test results (A/A-ST). The close alignment between A/B and A/B-ST columns, particularly for statistically significant effects (marked with *), demonstrates the method's ability to detect true treatment effects. The A/A-ST values statistically near zero confirm that the method avoids false positive detections. Each experiment uses the best-performing prediction model selected through cross-validation.}
\label{Tab:validation_histlabs}
\begin{tabular}[t]{lcccc}
\toprule
Experiment ID & A/B &  A/B-ST & A/A-ST & Selected Prediction Model \\
\midrule
{Experiment A}
& -0.14   & -0.28   & 0.13    & PCRRidge-2 \\
{Experiment B}
&  $-1.84^*$ & $-1.51^*$   & -0.45  & PCRRidge-2\\
{Experiment C}
&  $0.17^*$ & $0.31^*$   & -0.15  & PCRRidge-2\\
{Experiment D}
&  -0.45  &  -0.74   & 0.28  & PCRLasso-2\\
{Experiment E}
&  $-1.48^*$  &  $-1.36^*$   & -0.2  & Ridge-2\\
{Experiment F}
&  $0.22^*$  &  $0.15^*$   & 0.07  & PCRRidge-2\\
\bottomrule
\multicolumn{5}{l}{Notes. $^*$ indicates statistical significance at 0.05 level.} 
\end{tabular}
\end{table}

We observe that the ground truth ATE estimates using the actual controls are aligned with those of the counterfactual controls in terms of direction and statistical significance. Additionally, the A/A-ST tests did not indicate any statistically significant difference between the counterfactual and observed control outcomes at the 0.05 significance level. Hence, the two validations indicated above are successfully passed for these six experiments. This numerical study indicates that our counterfactual control outcomes from the two-phase method result in ATE estimates that are aligned with the ATEs computed using the observed control outcomes.

\subsection{Machine Learning Bias}

A more careful look at Table \ref{Tab:validation_histlabs} reveals a few concerns. Although the validations are passed, the magnitude of effects contains bias. In other words, the current analysis may be relied on for making directional launch or no-launch decisions. However, the magnitudes of the estimates are quite different, particularly in experiments B, C, D, and E, where the ATE is nonzero. This bias is due to the fact that the prediction models aimed to produce the most accurate predictions of the counterfactuals, which was achieved by producing biased estimates to reduce the variance. To better understand the implications of this bias for decision-making, we next examine how it evolves over time.

\subsubsection{Magnified Bias Over Time}\label{subsubsec:timedstudy}

A shortcoming of the above validation is that the training data for predicting the counterfactuals was data for time periods close to the experiment. However, if we want to test a launched treatment's performance over time (say, after several months), we have to use the dated pre-launch data to train the model. Given the dynamic nature of most data, the estimates may become highly biased and unreliable. To assess this, we conducted a time study using Experiment C where we trained the model on a 3-month-old dataset. The results are illustrated in Table \ref{Tab:validation_PCRRidge_2Ph_E1}.

\begin{table}[!htbp]
\centering
\caption{Impact of training data recency on treatment effect estimation. The table compares validation results using recent versus 3-month-old training data for Experiment C. The comparison demonstrates how the accuracy of counterfactual predictions degrades when using older training data, particularly evident in the increased bias of both A/B-ST estimates and the emergence of a false positive in the A/A-ST validation.}
\label{Tab:validation_PCRRidge_2Ph_E1}
\begin{tabular}[t]{lcccc}
\toprule
Experiment ID & A/B &  A/B-ST & A/A-ST & Prediction Model \\
\midrule
{Experiment C}& $0.17^*$   & $0.31^*$   & -0.15    & PCRRidge-2 \\
{Experiment C 3-month-old}&  $0.17^*$ & $0.42^*$   & {\color{pastelred}-$0.37^*$}  & PCRRidge-2\\
\bottomrule
\multicolumn{5}{l}{Notes. $^*$ indicates statistical significance at 0.05 level.} 
\end{tabular}
\end{table}

We observed that the ATE of the counterfactual control becomes more biased in the A/B-ST test as we use older training data. More importantly, the bias is no longer benign in the A/A-ST test and leads to a failed validation, as it detects an effect that does not exist.

Motivated by this observation, the need for a de-biasing mechanism becomes necessary to make the whole approach usable in this scenario. This is our next aim.

\subsection{Debiasing the Estimates} \label{sec:bias}

To further shrink the gap between the ATE estimates from the A/B experiments and the two-phase approach, we modified the hyperparameter tuning of the models in the second phase. Specifically, given the importance of bias, we combined the objective function of relative error, defined in Eq. \eqref{eq:relative-error}, with the bias. The new hyperparameter tuning objective function is defined by
\begin{equation}\label{eq:new-loss}
    \text{loss} = \text{relative error} + \alpha |\text{bias}|\,, 
\end{equation}
where $\alpha$ is a constant and the bias term is defined by
\[
\text{bias}=\frac{\sum_{it} (\text{Actual}_{it}-\text{Prediction}_{it})}{\sum_{it} 1}\,.
\]
We then used the new loss function from Eq. \eqref{eq:new-loss} in all model selection and hyperparameter tuning procedures. In our experiments, we found that the best results are achieved by setting $\alpha\approx 20$.

\paragraph{Alternative de-biasing approach.} We also experimented with another de-biasing approach based on sample-splitting techniques. Specifically, we split the donor set into two groups, using one for training and the other as a hold-out set to estimate the bias, and then used that to de-bias the model predictions. We implemented this for the Ridge-2 method, denoting the resulting technique as RidgeDebiased-2. Overall, we found this approach to be less effective than the above method.

\begin{figure}[!h]
    \centering
    \includegraphics[width=.7\textwidth]{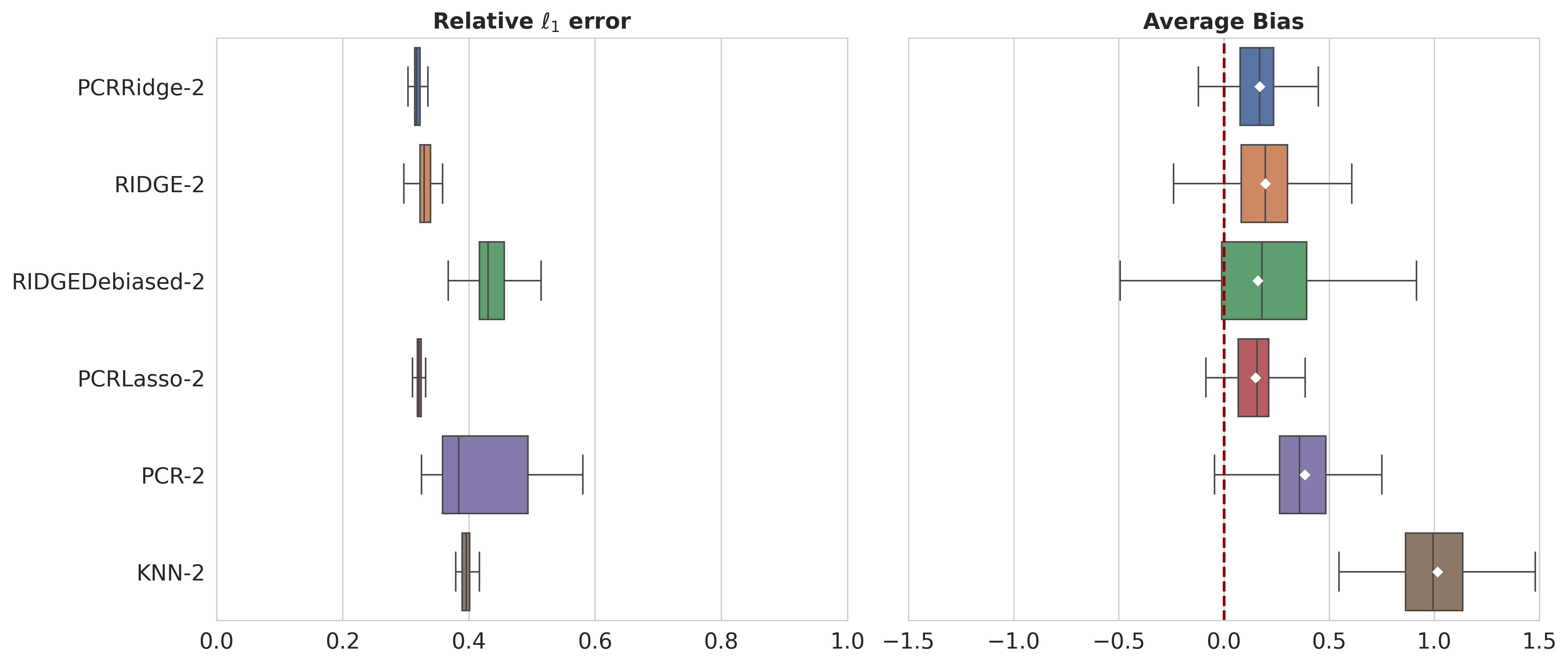}
    \caption{Comparison of prediction error (left) and bias distribution (right) across methods in Experiment C. PCRRidge-2, Ridge-2, and PCRLasso-2 implement debiasing through modified hyperparameter tuning, whereas RidgeDebiased-2 uses an alternative approach based on sample splitting. White dots in the right panel indicate distribution means.}
    \label{fig:bias-correction-experiment-C}
\end{figure}

\paragraph{Results of de-biasing.}
The relative error and average bias, after implementing the new parameter tuning approach as well as for RidgeDebiased-2 for Experiment C, are shown in Figure \ref{fig:bias-correction-experiment-C}. This shows that the methodology has been effective in reducing the bias of PCRRidge-2, RIDGE-2, RidgeDebiased-2, and PCRLasso-2. 

We next study how the best of these models perform on the experiments we studied before. Specifically, we focus on experiments where there was significant evidence of bias—i.e., a substantial gap between the ATE estimates from the A/B test and the two-phase approach, with non-overlapping confidence intervals. Table \ref{tab:biasReduction} summarizes the results, showing that the debiasing method reduced bias in the four experiments where bias was a concern.
\begin{table}[!h]
\centering
\caption{Effectiveness of the debiasing approach across historical experiments. The table presents results for four experiments where bias was initially a concern, showing how the debiased two-phase method produces estimates (A/B-ST) that closely match the experimental ground truth (A/B). The A/A-ST values near zero demonstrate that the debiasing successfully prevents false positive detections, while maintaining the ability to detect true treatment effects of varying magnitudes and directions.}
\label{tab:biasReduction}
\begin{tabular}[t]{lcccc}
\toprule
Experiment ID & A/B &  A/B-ST & A/A-ST & Prediction Model \\
\midrule
{Experiment B}
&  $-1.84^*$ & $-1.77^*$   & $-0.00$  & PCRLasso-2\\
{Experiment C}
&  $0.17^*$ & $0.20^*$   & $-0.02$  & PCRRidge-2\\
{Experiment E}
&  $-1.48^*$  &  $-1.43^*$   & $-0.04$  & PCRRidge-2\\
{Experiment F}
&  $0.22^*$  &  $0.15^*$   & $0.07$  & PCRRidge-2\\
\bottomrule
\multicolumn{5}{l}{Notes. $^*$ indicates statistical significance at 0.05 level.} 
\end{tabular}
\end{table}

Furthermore, we repeated the time study from Section \ref{subsubsec:timedstudy} using the de-biasing technique. The results, shown in Table \ref{Tab:validation_PCRRidge_2Ph_E1_after_debiasing} below, demonstrate that the biases are substantially reduced. More importantly, the failed validation of A/A-ST using the 3-month-old data no longer occurs.

\begin{table}[!h]
\centering
\caption{Effectiveness of debiasing in maintaining estimate accuracy over time. The table demonstrates how the debiased estimator resolves the temporal stability issues previously observed in Experiment C. Using both recent and 3-month-old training data, the estimates (A/B-ST) closely match the experimental ground truth (A/B), while the (A/A-ST) value remains near zero, indicating no false positive detections. This shows that the debiasing approach successfully maintains accuracy even when using historical training data.}
\label{Tab:validation_PCRRidge_2Ph_E1_after_debiasing}
\begin{tabular}[t]{lcccc}
\toprule
Experiment ID & A/B &  A/B-ST & A/A-ST & Prediction Model \\
\midrule
{Experiment C}& $0.17^*$   & $0.20^*$   & $-0.02$    & PCRRidge-2 \\
{Experiment C 3-month-old}&  $0.17^*$ & $0.16^*$   & $0.00$  & PCRRidge-2\\
\bottomrule
\multicolumn{5}{l}{Notes. $^*$ indicates statistical significance at 0.05 level.} 
\end{tabular}
\end{table}

       \section{Heterogeneous Treatment Effect (HTE) Estimation}\label{sec:HTE}

\begin{figure}[htb]
    \centering
    \includegraphics[width=.3\textwidth]{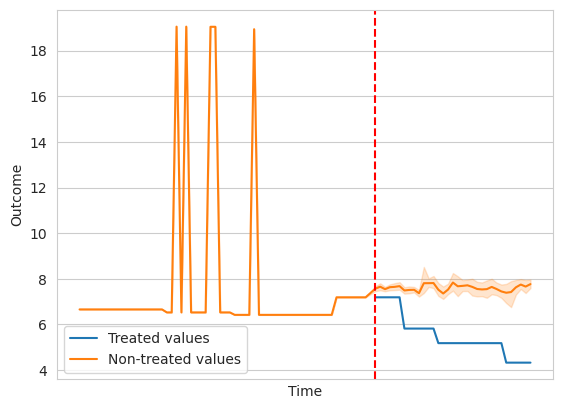}
\includegraphics[width=.3\textwidth]{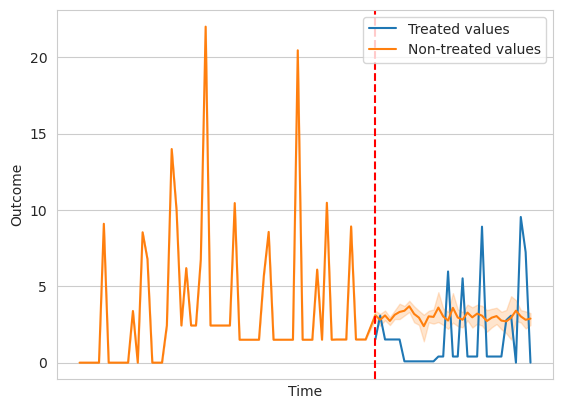} 
\includegraphics[width=.3\textwidth]{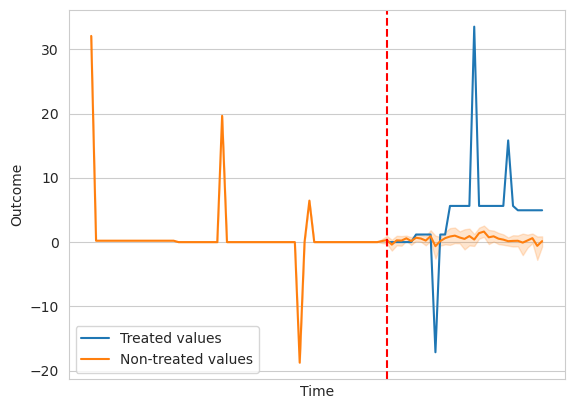}
    \caption{Heterogeneous treatment effects illustrated through product-level time series. Each panel shows outcomes for a different product: treated values (blue line) and non-treated values (orange line) over time, with the vertical red dashed line indicating treatment start. The orange line after treatment represents predicted counterfactual outcomes, while the blue line shows actual outcomes. The distinct patterns across products—showing both positive and negative gaps between treated and counterfactual values—demonstrate how treatment effects can vary substantially across units. Note the different scales and volatility patterns across products, highlighting the challenge of accurate counterfactual prediction in highly noisy and heterogeneous settings.}
    \label{fig:HTE}
\end{figure}

The proposed method can allow for counterfactual predictions of control outcomes at granular unit-time level, as also demonstrated by prior literature, e.g., \citep{agarwal2020synthetic}. This means that one can easily estimate the treatment effect for each unit identifying which units contribute most to the observed average effect. Figure \ref{fig:HTE} illustrates such predictions for three units or products (each panel represents predictions for a given product).

\section{Conclusion}
This study presents a reliable approach for measuring treatment effects when a control group is unavailable and data is highly noisy. Our method constructs counterfactual control outcomes by leveraging a large-scale set of donor units combined with prediction models designed for high-dimensional data. We validated this approach using six historical experiments where actual control outcomes were available, allowing us to assess the accuracy of our counterfactual predictions.

The key step in our approach is a two-phase methodology that applies nearest-neighbor matching before implementing prediction methods. This approach offers several advantages over traditional single-phase methods: it selects a more representative donor pool, reduces prediction error, and successfully replicates experimental results by both detecting true effects and avoiding false positives. Furthermore, this approach enables efficient use of unit covariate information in settings with large donor pools.

Our comprehensive assessment revealed an important challenge: while the two-phase approach accurately detects treatment effects and their direction, machine learning methods can introduce bias in effect size estimation. This bias is particularly evident when using historical training data for long-term treatment evaluation. To address this challenge, we developed a modified hyperparameter tuning approach that explicitly accounts for bias. This modification produces more accurate effect size estimates that maintain their reliability even when using historical training data.

\bibliographystyle{ormsv080}
\bibliography{references}

\begin{thebibliography}{19}
\expandafter\ifx\csname natexlab\endcsname\relax\def\natexlab#1{#1}\fi
\expandafter\ifx\csname url\endcsname\relax
  \def\url#1{{\tt #1}}\fi
\expandafter\ifx\csname urlprefix\endcsname\relax\def\urlprefix{URL }\fi
\expandafter\ifx\csname urlstyle\endcsname\relax
  \expandafter\ifx\csname doi\endcsname\relax
  \def\doi#1{doi:\discretionary{}{}{}#1}\fi \else
  \expandafter\ifx\csname doi\endcsname\relax
  \def\doi{doi:\discretionary{}{}{}\begingroup \urlstyle{rm}\Url}\fi \fi

\bibitem[{Abadie et~al.(2010)Abadie, Diamond, and
  Hainmueller}]{abadie2010synthetic}
Abadie, Alberto, Alexis Diamond, Jens Hainmueller. 2010.
\newblock Synthetic control methods for comparative case studies: Estimating
  the effect of california’s tobacco control program.
\newblock {\it Journal of the American statistical Association\/} {\bf
  105}(490) 493--505.

\bibitem[{Abadie et~al.(2015)Abadie, Diamond, and
  Hainmueller}]{abadie2015comparative}
Abadie, Alberto, Alexis Diamond, Jens Hainmueller. 2015.
\newblock Comparative politics and the synthetic control method.
\newblock {\it American Journal of Political Science\/} {\bf 59}(2) 495--510.

\bibitem[{Abadie and Gardeazabal(2003)}]{abadie2003economic}
Abadie, Alberto, Javier Gardeazabal. 2003.
\newblock The economic costs of conflict: A case study of the basque country.
\newblock {\it American economic review\/} {\bf 93}(1) 113--132.

\bibitem[{Agarwal et~al.(2020)Agarwal, Shah, and Shen}]{agarwal2020synthetic}
Agarwal, Anish, Devavrat Shah, Dennis Shen. 2020.
\newblock Synthetic a/b testing using synthetic interventions.
\newblock {\it arXiv preprint arXiv:2006.07691\/} .

\bibitem[{Amjad et~al.(2018)Amjad, Shah, and Shen}]{amjad2018robust}
Amjad, Muhammad, Devavrat Shah, Dennis Shen. 2018.
\newblock Robust synthetic control.
\newblock {\it The Journal of Machine Learning Research\/} {\bf 19}(1)
  802--852.

\bibitem[{{Arkhangelsky} and {Imbens}(2023)}]{arkhangelsky2023causal}
{Arkhangelsky}, Dmitry, Guido {Imbens}. 2023.
\newblock {Causal Models for Longitudinal and Panel Data: A Survey}.
\newblock {\it arXiv e-prints\/} \doi{10.48550/arXiv.2311.15458}.

\bibitem[{Athey et~al.(2021)Athey, Bayati, Doudchenko, Imbens, and
  Khosravi}]{athey2021matrix}
Athey, Susan, Mohsen Bayati, Nikolay Doudchenko, Guido Imbens, Khashayar
  Khosravi. 2021.
\newblock Matrix completion methods for causal panel data models.
\newblock {\it Journal of the American Statistical Association\/} {\bf
  116}(536) 1716--1730.

\bibitem[{{Athey} and {Imbens}(2016)}]{athey2016economics}
{Athey}, Susan, Guido {Imbens}. 2016.
\newblock {The Econometrics of Randomized Experiments}.
\newblock {\it arXiv e-prints\/} \doi{10.48550/arXiv.1607.00698}.

\bibitem[{Blake et~al.(2021)Blake, Moshary, Sweeney, and
  Tadelis}]{blake2021price}
Blake, Tom, Sarah Moshary, Kane Sweeney, Steve Tadelis. 2021.
\newblock Price salience and product choice.
\newblock {\it Marketing Science\/} {\bf 40}(4) 619--636.
\newblock \doi{10.1287/mksc.2020.1261}.

\bibitem[{Bottmer et~al.(2024)Bottmer, Imbens, Spiess, and
  Warnick}]{bottmer2024designbased}
Bottmer, Lea, Guido~W. Imbens, Jann Spiess, Merrill Warnick. 2024.
\newblock A design-based perspective on synthetic control methods.
\newblock {\it Journal of Business \& Economic Statistics\/} {\bf 42}(2)
  762--773.

\bibitem[{Chen and Donoho(1995)}]{chen1995examples}
Chen, Scott~S., David~L. Donoho. 1995.
\newblock Examples of basis pursuit.
\newblock {\it Proceedings of Wavelet Applications in Signal and Image
  Processing III\/}. San Diego, CA.

\bibitem[{Cox(1958)}]{cox1958planning}
Cox, David~Roxbee. 1958.
\newblock {\it Planning of experiments.\/}.
\newblock Wiley.

\bibitem[{Doudchenko and Imbens(2017)}]{imbens2017balancing}
Doudchenko, Nikolay, Guido~W. Imbens. 2017.
\newblock Balancing, regression, difference-in-differences and synthetic
  control methods: A synthesis .

\bibitem[{Farias et~al.(2021)Farias, Li, and Peng}]{farias2021learning}
Farias, Vivek, Andrew~A Li, Tianyi Peng. 2021.
\newblock Learning treatment effects in panels with general intervention
  patterns.
\newblock A.~Beygelzimer, Y.~Dauphin, P.~Liang, J.~Wortman Vaughan, eds., {\it
  Advances in Neural Information Processing Systems\/}.

\bibitem[{Gavish and Donoho(2014)}]{gavish2014optimal}
Gavish, Matan, David~L Donoho. 2014.
\newblock The optimal hard threshold for singular values is $4\sqrt{3}$.
\newblock {\it IEEE Transactions on Information Theory\/} {\bf 60}(8)
  5040--5053.

\bibitem[{Gupta et~al.(2019)Gupta, Kohavi, Tang, Xu, Andersen, Bakshy, Cardin,
  Chandran, Chen, Coey, Curtis, Deng, Duan, Forbes, Frasca, Guy, Imbens,
  Saint~Jacques, Kantawala, Katsev, Katzwer, Konutgan, Kunakova, Lee, Lee, Liu,
  McQueen, Najmi, Smith, Trehan, Vermeer, Walker, Wong, and
  Yashkov}]{gupta2019top}
Gupta, Somit, Ronny Kohavi, Diane Tang, Ya~Xu, Reid Andersen, Eytan Bakshy,
  Niall Cardin, Sumita Chandran, Nanyu Chen, Dominic Coey, Mike Curtis, Alex
  Deng, Weitao Duan, Peter Forbes, Brian Frasca, Tommy Guy, Guido~W. Imbens,
  Guillaume Saint~Jacques, Pranav Kantawala, Ilya Katsev, Moshe Katzwer, Mikael
  Konutgan, Elena Kunakova, Minyong Lee, MJ~Lee, Joseph Liu, James McQueen,
  Amir Najmi, Brent Smith, Vivek Trehan, Lukas Vermeer, Toby Walker, Jeffrey
  Wong, Igor Yashkov. 2019.
\newblock Top challenges from the first practical online controlled experiments
  summit.
\newblock {\it SIGKDD Explor. Newsl.\/} {\bf 21}(1) 20–35.

\bibitem[{Kohavi et~al.(2007)Kohavi, Henne, and
  Sommerfield}]{kohavi2007practical}
Kohavi, Ron, Randal~M Henne, Dan Sommerfield. 2007.
\newblock Practical guide to controlled experiments on the web: Listen to your
  customers not to the hippo.
\newblock {\it Proceedings of the 13th ACM SIGKDD International Conference on
  Knowledge Discovery and Data Mining\/}. ACM, 959--967.

\bibitem[{Rubin(1978)}]{rubin1978bayesian}
Rubin, Donald~B. 1978.
\newblock Bayesian inference for causal effects: The role of randomization.
\newblock {\it The Annals of statistics\/}  34--58.

\bibitem[{Tibshirani(1996)}]{tibshirani1996regression}
Tibshirani, Robert. 1996.
\newblock Regression shrinkage and selection with the lasso.
\newblock {\it Journal of the Royal Statistical Society Series B\/} {\bf 58}
  267--288.

\end{thebibliography}

\end{document}